\documentclass[letterpaper,10pt,conference]{ieeeconf}  

\IEEEoverridecommandlockouts                              

\usepackage{graphicx}
 \usepackage{algorithm}
\usepackage{algorithmicx}
\usepackage[noend]{algpseudocode}
\usepackage{amsmath}
\usepackage{amsfonts}
\usepackage{wasysym}
\usepackage{diagbox}
\usepackage[colorlinks,
linkcolor=blue,
urlcolor=blue,
anchorcolor=blue, 
citecolor=blue, 
]{hyperref}
\usepackage{xcolor}
\usepackage{caption}
\usepackage{subfigure}
\usepackage{cuted}
\usepackage{adjustbox}
\usepackage{rotating}
\usepackage[normalem]{ulem}

\allowdisplaybreaks[4]
\usepackage{balance}

\title{\LARGE \bf
From ``Thumbs Up'' to ``10 out of 10'':  Reconsidering Scalar Feedback in Interactive Reinforcement Learning

}

\author{Hang Yu$^{1}$, Reuben M. Aronson$^{1}$, Katherine H. Allen$^{2}$,  and Elaine Schaertl Short$^{1}$
\thanks{*The work described herein was funded in part by the Henry Luce Foundation Clare Booth Luce Fellowship Program and the
US National Science Foundation (IIS 2132887).}
\thanks{$^{1}$Hang Yu, Reuben M. Aronson, and Elaine Schartl Short are with Tufts University School of Engineering, Computer Science. Medford, Massachusetts, United States of America
        {\tt\small \{hang.yu625917, reuben.aronson, elaine.short\}@tufts.edu}}%
\thanks{$^{2}$Katherine H. Allen is with Tufts University School of Engineering, Mechanical Engineering. Medford, Massachusetts, United States of America
        {\tt\small kat.allen@tufts.edu}}%
}

\begin{document}

\maketitle
\thispagestyle{empty}
\pagestyle{empty}

\begin{abstract}
Learning from human feedback is an effective way to improve robotic learning in exploration-heavy tasks. 
Compared to the wide application of binary human feedback, 
scalar human feedback has been used less because it is believed to be noisy and unstable.
In this paper, 
we compare scalar and binary feedback, and demonstrate that scalar feedback benefits learning when properly handled. 
We collected binary or scalar feedback respectively from two groups of crowdworkers on a robot task. 
We found that when considering how consistently a participant labeled the same data, scalar feedback led to less consistency than binary feedback; however, the difference vanishes if small mismatches are allowed. 
Additionally, scalar and binary feedback show no significant differences in their correlations with key Reinforcement Learning targets. 
We then introduce Stabilizing TEacher Assessment DYnamics (STEADY) to improve learning from scalar feedback.
Based on the idea that scalar feedback is muti-distributional, 
STEADY re-constructs underlying positive and negative feedback distributions and re-scales scalar feedback based on feedback statistics.
We show that models trained with \textit{scalar feedback + STEADY } outperform baselines, including binary feedback and raw scalar feedback, in a robot reaching task with non-expert human feedback.
Our results show that 
both binary feedback and scalar feedback are dynamic, and 
scalar feedback is a promising signal for use in interactive Reinforcement Learning.

\end{abstract}

\section{Introduction}

Interactive Reinforcement Learning is a method that reduces data needs and improves learning efficiency by having a human-in-the-loop providing \emph{evaluative feedback} to an agent during learning.
Evaluative feedback is natural for non-experts to provide, and can take the form of either \emph{binary} feedback, in which feedback can only be either ``good'' or ``bad'', or \emph{scalar} feedback, which takes a value from a scale of values (e.g., ``1-10'', ``A-F'', or ``0-5 stars''). In theory, both of these types of feedback can contain useful information.  However, only binary feedback has been widely used in prior work\cite{knox2008tamer,griffith2013policy,tan2020top}
%
since it is easy to separate into positive and negative, and limiting people to only two options can reduce noise.

Despite these advantages,
allowing \emph{only} binary feedback reduces information in the signal such as the intensity of preferences. 
Scalar feedback, on the other hand, is also naturally used in daily life (e.g., movie rating and product evaluation) but encodes information beyond that in binary feedback, such as ranking and magnitude, through a wider range of possible values.  However, scalar feedback is more difficult to apply to RL; it is likely to have minor differences while evaluating the same data and is more difficult to separate into good and bad.  Furthermore, users are unlikely to scale their feedback such that it can be interpreted directly as a reward.  For example, on a scale of 10, the distance from 5 to 7 may be perceived as different from the distance from 3 to 5 although they are numerically the same.
These difficulties have led to prior work categorizing scalar feedback as non-optimal
\cite{arzate2020survey,https://doi.org/10.48550/arxiv.1811.07871}
and have led researchers to underestimate the potential of scalar feedback in interactive RL. 



\begin{figure}[t]

    \centering
    \includegraphics[width = 0.65\linewidth]{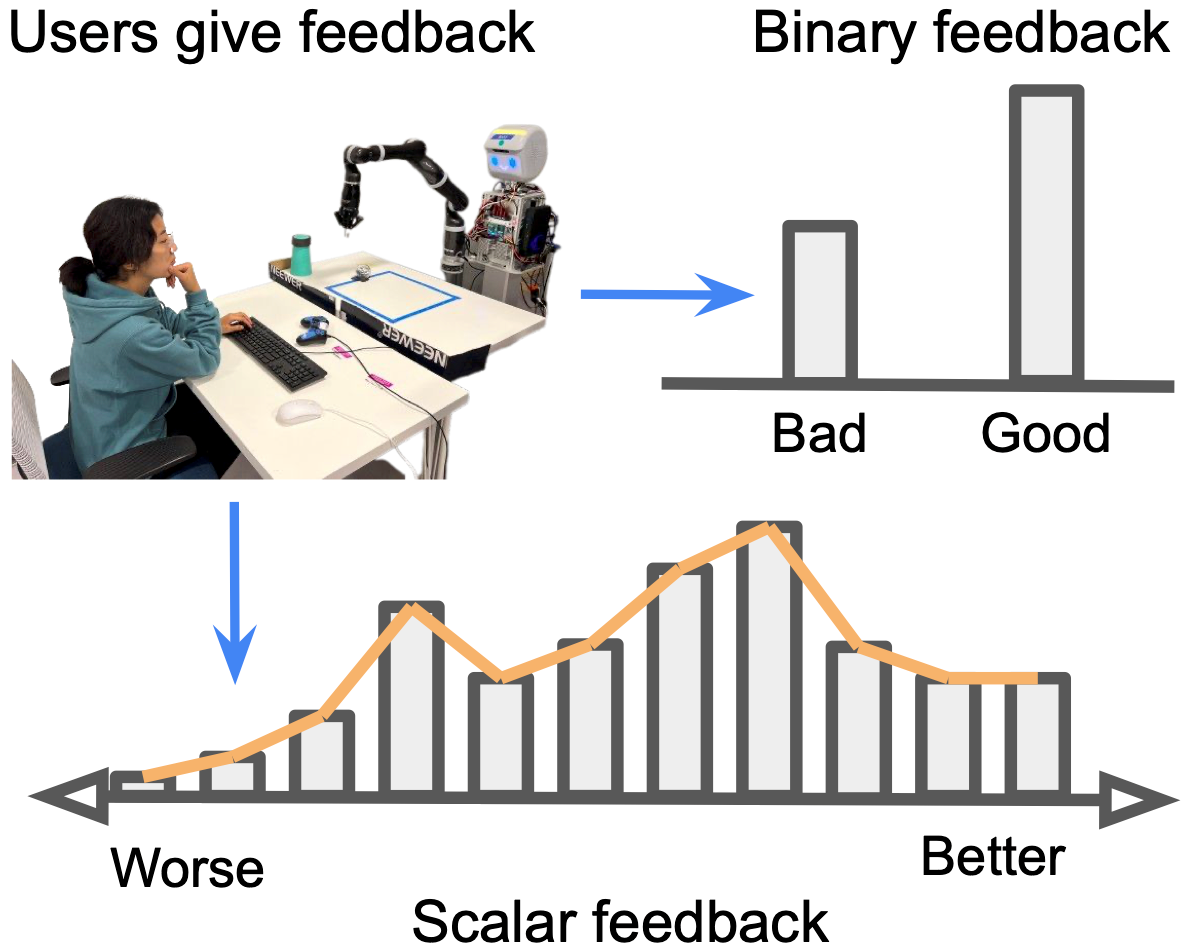}
    \label{fig:page1}
    \caption{ 
    Binary feedback is strictly separated into ``good'' and ``bad'' and thus  is noise reduced.
    Scalar feedback allows users to express preferences via a larger scale of values. 
    }
    \vspace{-0.9em}
\end{figure}
In this work, we conduct a study of scalar and binary feedback with 90 online participants evaluating a robot performing a manipulation task, and show that scalar feedback has the potential to improve learning when properly handled. 
We show that human teachers can be inconsistent when giving both scalar and binary feedback even in a task (reaching to and pressing a button) that is among the most easy-to-understand tasks a robot might do.
We study the correlation of scalar and binary feedback with interactive RL training targets, and fail to find a significant difference between scalar and binary feedback, although we find that scalar feedback has more variation. 
Given that these results suggest that scalar feedback is not necessarily worse for learning if we can address the scaling problem \cite{ibarz2018reward} and noise,
we present Stabilizing TEacher Assessment DYnamics (STEADY), an unsupervised feedback filter.
The key insight of STEADY is to use a multi-distributional model for scalar feedback,  re-normalizing noisy scalar feedback to confidence scores (i.e. more precise magnitudes) and recovering positive/negative labels.
This enables binary-feedback-based learning algorithms to learn
effectively from information-rich scalar feedback and results in a significant improvement over binary feedback. 
We demonstrate that models trained with \textit{scalar feedback + STEADY} outperform models trained with binary feedback on a robot learning task.
 
\begin{figure*}[t]
  \centering
  \setlength{\abovecaptionskip}{0.2cm}
    \includegraphics[width=1\linewidth]{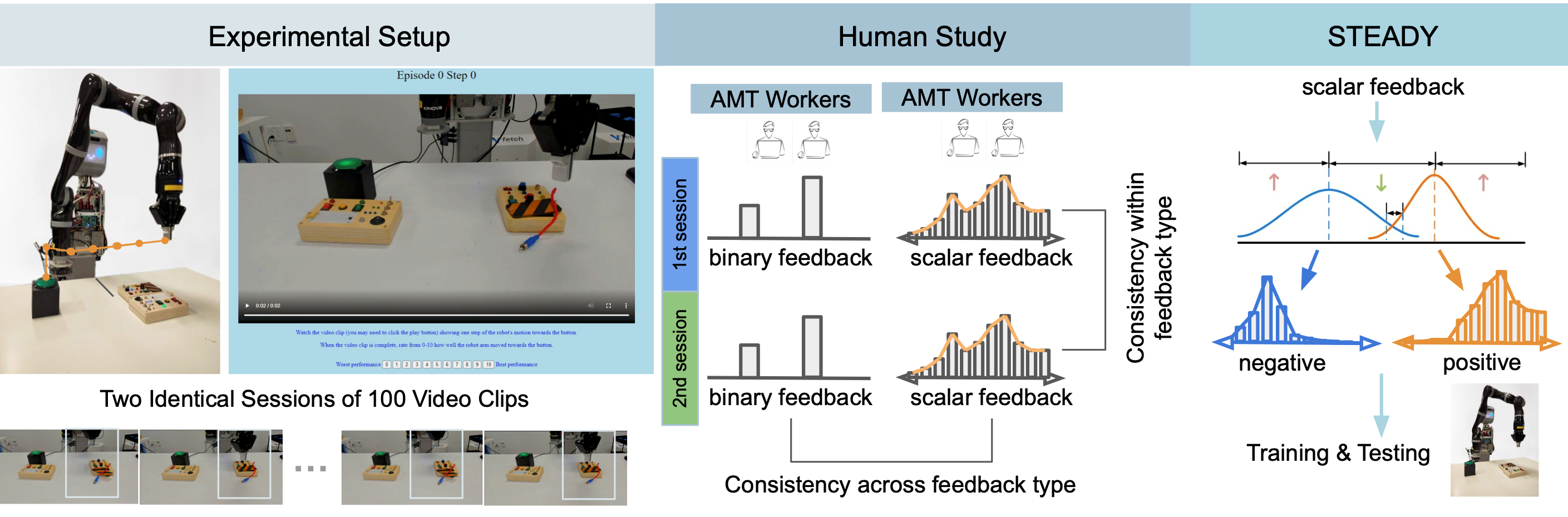}
  \caption{Overview. We recorded two identical sessions
of 100 video clips of a robot performing a button-pressing task.
  90 online workers were recruited, with one group providing binary feedback and the other group providing scalar feedback.
   We investigated feedback dynamics by comparing feedback in the two sessions within individuals and feedback between participants. 
  We validated that \textit{scalar feedback + STEADY} outperforms binary feedback on the button-pressing task. 
  }
  \label{fig:bp} 
    \vspace{-0.9em}
\end{figure*}
To the best of our knowledge,
this is the first work that quantifies feedback dynamics in binary feedback and scalar feedback,
indicating that feedback dynamics within individuals and between participants should be taken into consideration while using both binary feedback and scalar feedback.
STEADY is the first algorithm that enables binary-feedback-based algorithms to learn from 
scalar rewards 
 even when those rewards do not have a clear division into positive and negative. 
This work bridges the research of learning from binary feedback and scalar feedback, and demonstrates the potential of scalar feedback as a teaching signal.

\section{Background}

\label{sec:RW}

Interactive information in various forms from humans is widely used in prior work: 
as reward or feedback to shape learned policies~\cite{knox2008tamer,knox2009interactively, thomaz2005real,cederborg2015policy,griffith2013policy, yu2021active};
as signals, such as state visiting counts, prediction errors, and actions-states, to perform inverse reinforcement learning \cite{gregor2016variational,pathak2017curiosity,bellemare2016unifying,wang2020reward};
as demonstrations to enable imitation learning or behavior cloning \cite{hester2017deep, ho2016generative}; and
as preferences, which indicate preferable actions or trajectories \cite{christiano2017deep, ibarz2018reward}.
Three representative methods are Policy Shaping \cite{cederborg2015policy, griffith2013policy}, TAMER \cite{knox2008tamer, knox2009interactively}, and COACH \cite{pmlr-v70-macglashan17a}. 
In policy shaping, people give binary feedback on a robot's behaviors to indicate whether it is correct and thus shape the policies learned from environmental rewards.
In TAMER, a human's feedback is used as a reward signal to train a policy. 
COACH also uses human feedback as a reward signal, but instead of direct rewards, COACH replaces \emph{advantaged rewards} with human feedback, arguing that scalar feedback better matches the advantage function than the reward function.
Although TAMER and COACH theoretically can use scalar feedback,
they have primarily been evaluated with binary feedback or trinary feedback (e.g. -1, +1, +4 in \cite{pmlr-v70-macglashan17a}).
According to \cite{faulkner2020interactive}, the accuracy of human feedback can significantly impact learning efficiency with algorithms that expect binary feedback.
Throughout the body of prior work, there is a strong focus on using binary feedback to improve learning, with limited exploration of scalar feedback. 
 Our work provides the research community with a more explicit understanding of the differences between scalar and binary feedback and proposes a method that enables existing binary-feedback interactive RL algorithms to use scalar feedback.


Another area of related work characterizes the dynamics of how humans provide feedback to learning agents, both robots and other humans. 
For example, human reward signals correspond with both past actions and future rewards \cite{thomaz2006reinforcement}, the contingency of robot feedback has an impact on participants' tutoring behaviors \cite{fischer2013impact}, and 
temporal details are critical information for researchers to comprehend human expectations of joint attention
\cite{yu2010investigating}.
From the view of robots, research has shown that robots' action style \cite{zafari2019you} or deceptive feedback \cite{shim2016other} can affect participants. 
Education research has also shown that teacher feedback is not only related to the objective truth but also affected by human personalities, emotions, and characteristics, many of which may change over time and manifested through values (e.g. \cite{butler2007effect} and \cite{magill1994influence}). 

Moreover, teaching a robot not only concerns teaching behaviors, but also learning (since human teachers can learn to teach better), which has been shown to involve emotions that change over time \cite{picard2001toward}.
Furthermore, users learn to how to teach the robot during the interaction process \cite{kim2009people}, which may change their feedback as they become more familiar with the system. Thus, feedback is not a static value, even in what appears to the learning agent as the same state. Instead, feedback varies with individuals' emotions, personalities, and expectations about the future as well as the type and contingency of robots' behavior. Our work contributes to this literature with a direct comparison of feedback dynamics for binary and scalar feedback in a realistic robot reaching task and guidelines for using scalar feedback.

\section{Methodology}


We conducted an online user study to investigate binary and scalar feedback consistency over time and between users (\autoref{fig:bp}).
Two groups of participants were asked to evaluate the robot's performance while the robot was completing a button-pressing task.
The robot performed six trajectories twice, allowing us to detect changes in feedback by comparing the difference in feedback between the two sessions. All users saw the same trajectories, allowing us to detect variation in feedback between users on the same task with the same feedback type.
We showed how feedback correlates to the training targets of different algorithms. 
These results, and the data collected, motivate and enable the development and evaluation of our STEADY algorithm.

\subsection{Robot Task}
Participants evaluated the robot's performance on a button-pressing task. 
The robot is a humanoid robot with a 7-degree-of-freedom Kinova Jaco arm with a Robotiq gripper. %
A RealSense camera on the robot's head is used to identify the button's location and return the button's depth. 
In the task, the robot's goal is to navigate to the button's location and  
press the green button. 
The robot's observation is the current position of the gripper and the position of the button.
The robot's actions are going in one direction (left, right, backward, forward, or down).  
The move distance for each action is not deterministic (about 4 to 5 centimeters) due to noise in the manipulation pipeline of the robot.   

We used a relatively simple task to reduce the cognitive load on solving the task and thus reduce feedback dynamics from confusion over the task. 
While this is not a highly-complex task from the perspective of the robot learning literature, we used a relatively high-resolution representation (700 states, 3500 state-action pairs) and allowed the robot to have movement noise.
It is comparable to tasks used in interactive RL with real robots \cite{pmlr-v70-macglashan17a, kessler2019active, cui2020empathic, DBLP:journals/corr/abs-2110-11385}, where training time needs to be limited, and is easily understandable by non-expert users. 


\subsection{Experimental Setup}
\label{subsec: exp set}
During the study, participants were asked to watch the robot performing the task and to give evaluative feedback, either by clicking a button that said good or bad or by selecting a number between zero and ten.  
We did not use a range for scalar feedback that contains negative values since scales with no negative values are frequently used in daily life and including negative values naturally biases users toward separating scalar feedback into positive and negative, which would unfairly benefit scalar feedback in this experiment.
We generated the videos by recording the robot's behaviors in performing the task.
The robot's behavior was sampled from a partially-trained Q-Learning model.
The model is trained by objective rewards. 
The robot received 0.4 to 0.5 rewards (based on distance) if it moved toward the button,  -0.4 to -0.5 rewards if it moved away from the button, +10  if the robot successfully pressed the button, and -1 reward if it went down at the wrong location.
Videos were divided into clips. 
There were 200 video clips in total, each 3 seconds long (about 10 minutes total) with the first 100 and second 100 clips identical to each other; each clip represented one action performed by the robot; 
clips are in sequential order of a trajectory;
each 100 clips consisted of six trajectories (three successes, three failures).
No indication was given to participants that the two 100-clip ``sessions'' were the same.
To validate that participants would not notice the duplications,
we conducted a pilot study  by recruiting three robot experts from a robotic lab. We asked if they were aware that all clips were repeated once.
All of them answered \textit{no}.


\subsection{Online Participant Recruitment} 
The study was approved by the university review board and conducted online through 
Amazon Mechanical Turk (AMT). 
In order to reflect the feedback abilities of non-expert crowdworkers, we did not set worker requirements to only collect data from top workers with AMT Masters Qualification or workers with backgrounds. However, we did use filters to filter out click bots (HIT Approval Rate $\geq$ 95\% and Number of HITs Approved $\geq$ 50).
A total of 90 AMT crowdworkers were recruited, divided evenly between two groups (45 participants per group).
One group of workers was asked to give binary feedback and the other group of workers was asked to give scalar feedback. All workers received the same compensation and viewed the same content.

\subsection{Experimental Procedure}
Participants were redirected to the experiment website from the Amazon Mechanical Turk  (AMT) website after they clicked the \textit{accept \& work} button, where they viewed a welcome page and filled out a consent form to ensure that they were qualified for our study.  
After completing the consent documents, participants were shown how the robot performs the task by watching a short video and reading a short instruction to ensure that they had a consistent understanding of the task goals.
The demonstration video shows a human performing the task instead of a robot arm to avoid biasing their feedback to the robot.
After the demonstration video, the participant started the study.
They watched 200 video clips about the robot performing the button-pressing task. After watching one video clip, the participants gave feedback by clicking the button with the corresponding value, which caused the page to jump to the next clip.
The webpage displayed the \textit{thank you page} once all clips had been evaluated and provided them a validation code, ending the study.   
The validation code is used to redeem their compensations via the AMT platform.

\subsection{Hypotheses}

\noindent\textbf{H1. Human binary feedback changes less in the second session than scalar feedback} From prior work, we expect that binary feedback is noise-reduced and scalar feedback tends to be unstable, but none of the prior work has closely examined it.
To test this, we compare feedback between the first and second sessions with regards to self-agreement, feedback patterns, and bias in values. 


\noindent\textbf{H2. Human feedback patterns tend to be different.} 
We expect that human feedback patterns, especially scalar feedback patterns, are statistically significantly different from each other between users. The uniqueness of the feedback pattern explains why
learning from multiple people is a common issue in Interactive RL.

 \begin{figure*}[t]
    \centering
    \setlength{\abovecaptionskip}{0.2cm}
    \subfigure[Binary feedback]{
    \includegraphics[width=0.47\linewidth]{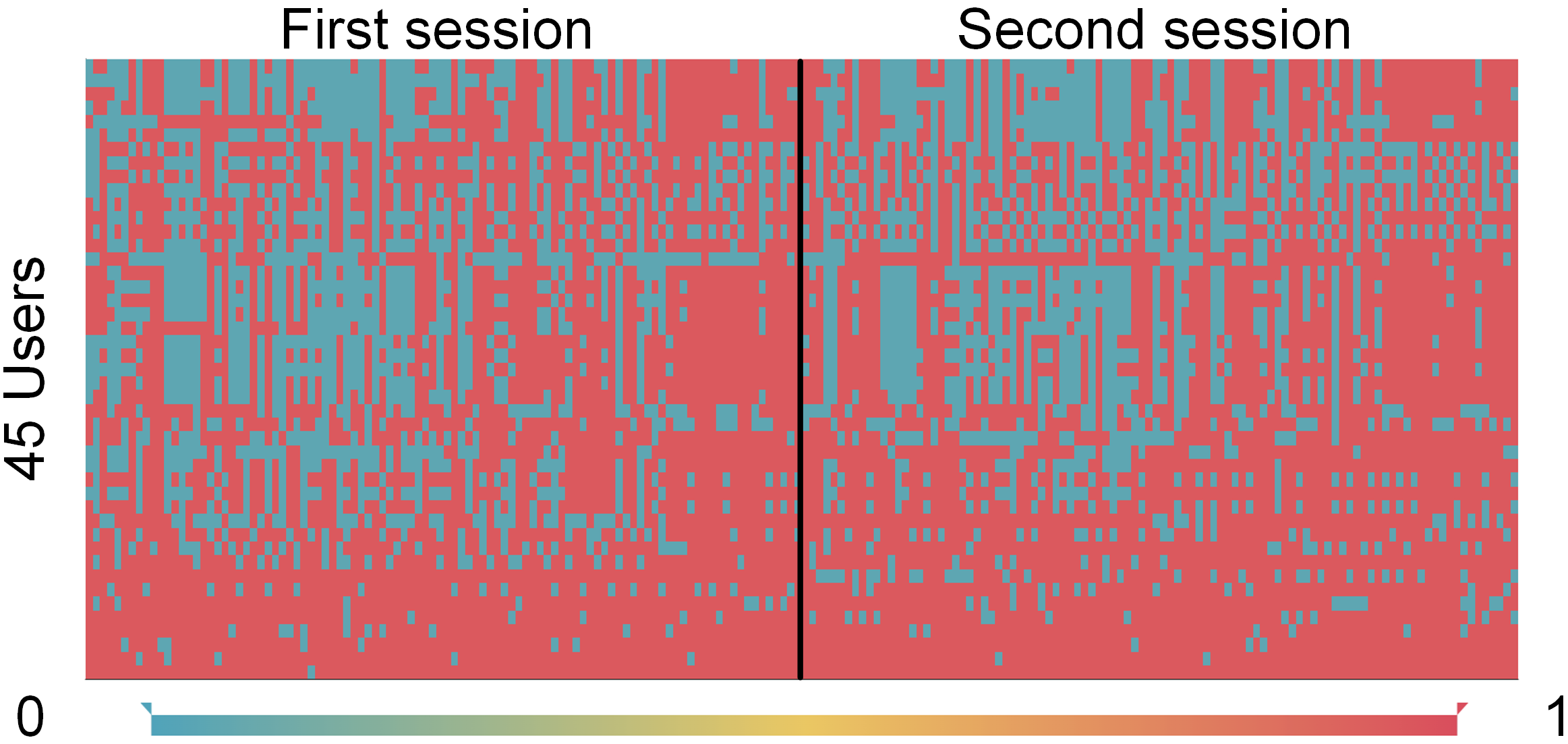}}
        \subfigure[Scalar feedback]{
    \includegraphics[width=0.47\linewidth]{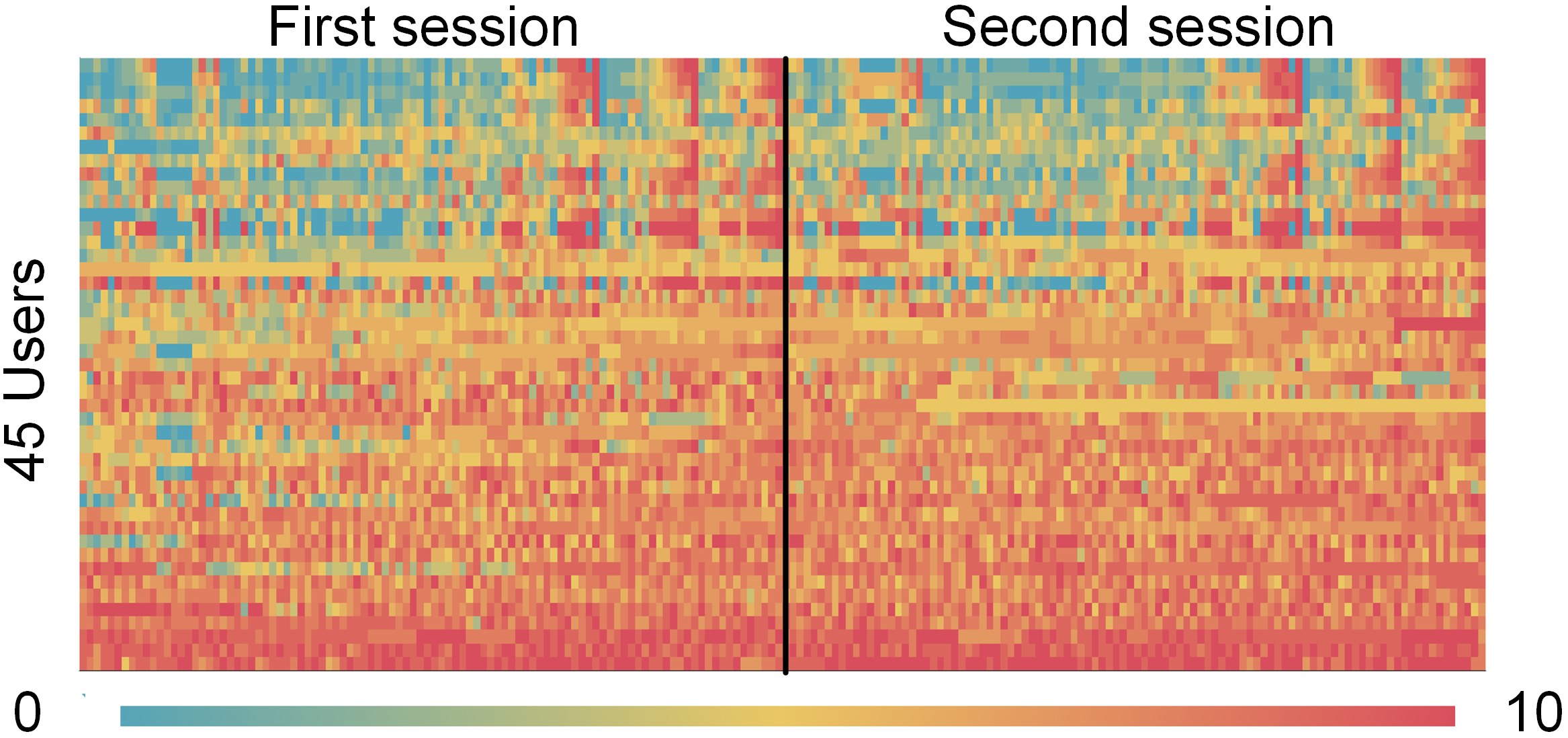}}
    \label{subfig:scale overview}
    \caption{Heatmaps of data collected from 90 participants. Each block shows a feedback, red is high-value and blue is low-value feedback. The x-axis represents the video clip and each row is a user, sorted by average feedback value. 
    Participants showed self-agreements but great differences
from each other. 
    }
    \label{fig:overview}
    \vspace{-0.7em}
\end{figure*}

\noindent\textbf{H3. Scalar feedback is less well correlated with a fully trained policy than binary feedback.} A significant concern with scalar feedback is that its noisiness and inconsistency decrease its usefulness for reinforcement learning. 





\section{User Study Results}


%
We show the overview of all feedback we collected during the human study in \autoref{fig:overview}.
Participants have diverse feedback patterns, but there are observable agreements in two sessions, and consistency between participants within the same group.  We investigate both the consistent and diverse aspects of feedback in this section. 
According to the average time per assignment provided by AMT, total study times on average were 29:22 minutes in the binary and 29:56 minutes in the scalar feedback condition. The study websites and the instructions are near-identical. This suggests that asking participants to give scalar feedback did not substantially increase the participant's effort, likely because scalar feedback is common in daily life. 


Using Shapiro-Wilk tests, we determined that the results do not from a normal distribution. Therefore, we applied Kruskal-Wallis H Tests and Wilcoxon Rank-Sum Tests to our results to test our hypotheses. 
For the scenarios that we performed Wilcoxon Rank-Sum Test multiple times, we used Holm-Sidak method to correct the p-values.   

\subsection{Analysis of H1: Did participants give consistent feedback in the two sessions?}
\label{subsection:fdwp}
\begin{figure}[t]

    \centering
    
    \includegraphics[width=0.85\linewidth]{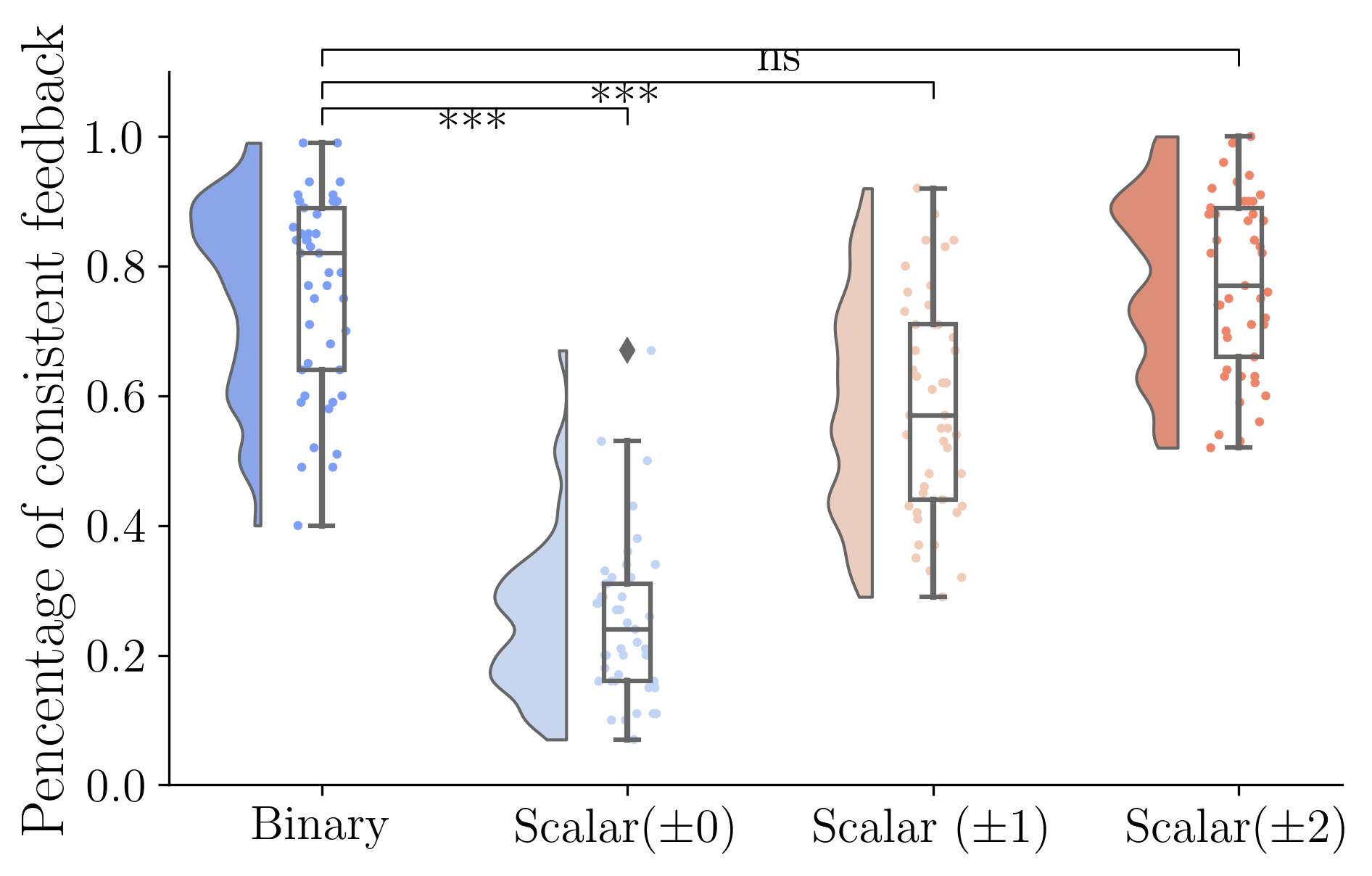}
    \caption{Self-Agreement.
    Each point represents the percentage of the same feedback in the first 100 and the second 100 feedback.
    Human feedback is not fully self-agreed. 
    Scalar feedback has similar self-agreement to binary feedback if a minor mismatch is allowed.
    }
    \vspace{-0.9em}
\label{fig:self-argree}
\end{figure}

\paragraph{Self-agreement on values.} 
To answer the question of whether participants are self-consistent,
we calculated the agreement of their first and second feedback sessions (\autoref{fig:self-argree}).
Each point in  \autoref{fig:self-argree} represents the percentage of self-agreed feedback for one participant.
Two scalar feedback values are considered to agree with each other if the difference between them is less than a threshold.
Using Kruskal-Wallis H Test, we found a significant difference in results $H = 111.5, p < 0.0001$. 
The average self-agreed feedback percentage rates over 45 participants are  76.3\% for binary feedback, 25.2\% for scalar feedback with a threshold of 0, 58.2\% with a threshold of 1, and 77.7\% with a threshold of 2. If exact matches are required, binary feedback is significantly more self-consistent than scalar feedback ($p < 0.0001$ for threshold $0$ and $\pm1$), which supports \textbf{H1}. However, when we increase the threshold of agreement, the self-consistency of scalar feedback improves, and by a threshold of $\pm 2$ the difference with binary feedback vanishes ($p = 0.775$).

\paragraph{Feedback pattern changed in the second session}
By comparing the averages in the two sessions, 
we found that most participants (74 out of 90) had a bias in their feedback in the second session as compared to the first.  We defined bias as a difference in the averages of the two sessions greater than 0.5 for scalar feedback or 0.05 for binary feedback (i.e. their average feedback changed by more than about 10\%).
More than half of the participants (23 binary, 25 scalar) were positively biased, while only 16 participants were non-biased (12 binary, 4 scalar).
This might be because the task is relatively simple and participants expected the robot to perform well on the task.
To study whether participants had significantly changed feedback-giving patterns during the interaction,
we performed a 
Wilcoxon Rank-Sum test over each participant's first 100 feedback and second 100 feedback, and corrected 
p-values by using the Holm-Sidak method. After correction,
15 out of 45 participants that gave binary feedback and 27 out of 45 participants that gave scalar feedback gave significantly different feedback ($p < 0.05$) in the two sessions.
This supports \textbf{H1}, and also 
 suggests that scalar feedback is more sensitive to changes in feedback.

\subsection{Analysis of H2: Were participants' feedback consistent with each other?}

\begin{figure}
    \centering
    \subfigure[Average feedback values]{
         \label{subfig:avgs}
    \includegraphics[width=0.46\linewidth]{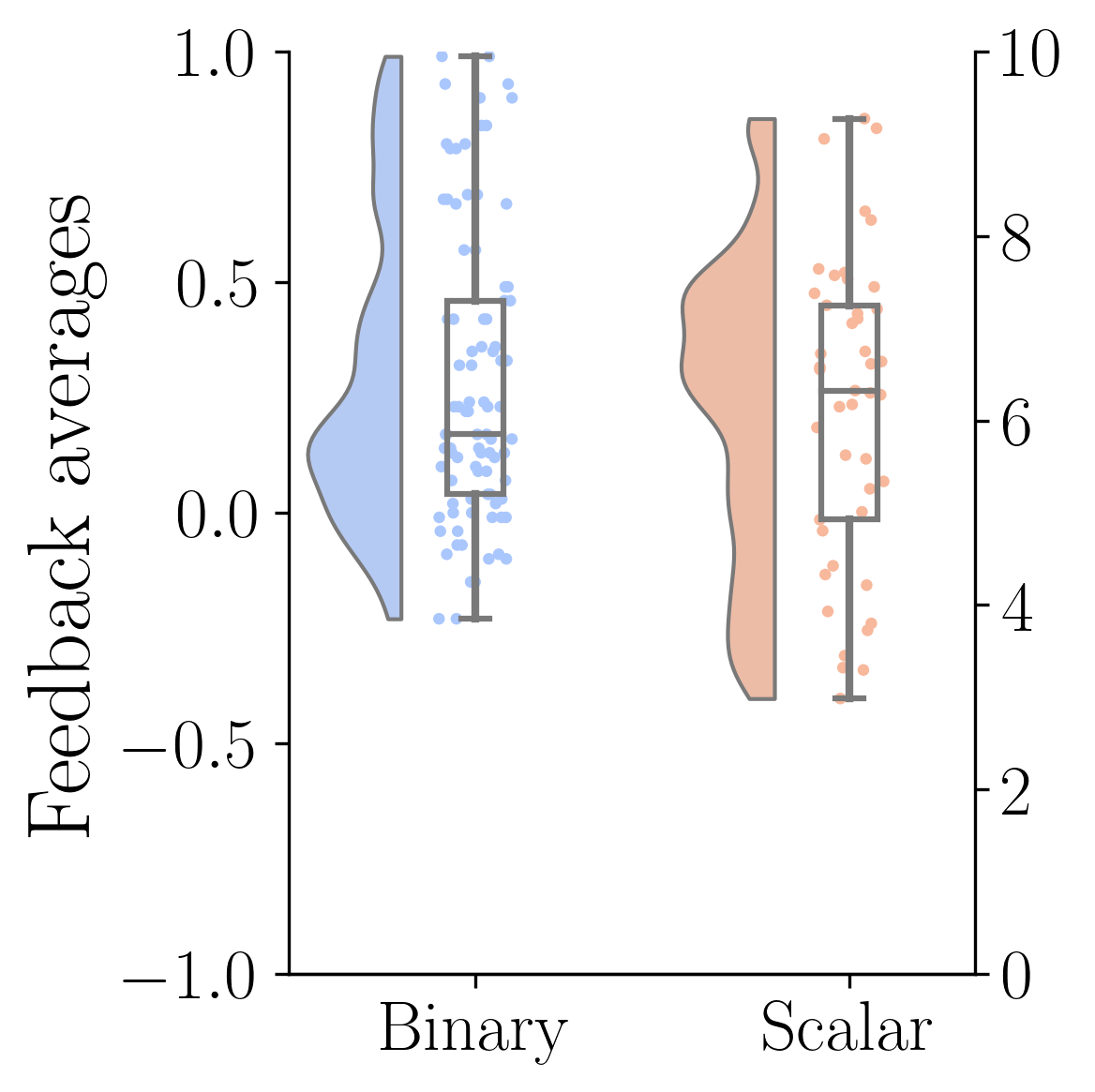}
    }
    \subfigure[Feedback pattern uniqueness]{
    \label{subfig:pttbp}
    \includegraphics[width=0.43\linewidth]{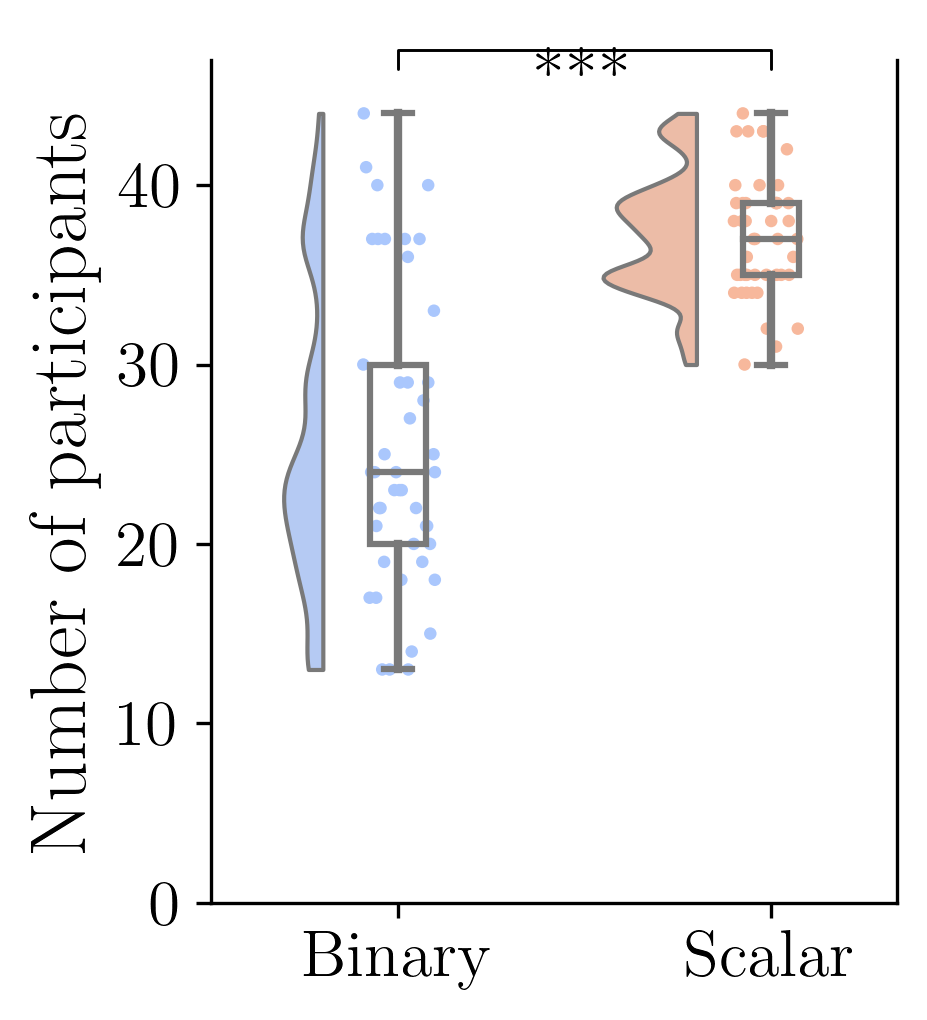}
    }

    \caption{Participants' feedback patterns deviate from each other: 
    A significant difference in the feedback pattern 
    and the feedback value widely exists.}
    \label{fig:dynamics among}
    \vspace{-0.9em}
\end{figure}

\paragraph{Participants rarely agreed with each other exactly}
Figure \autoref{subfig:avgs} shows the average feedback values of each participant, which cover a large range. 
Some participants preferred high-value feedback, while the majority gave balanced-value feedback, which is consistent with the conclusion in prior work~\cite{sidana2021user}. 
Using Kruskal-Wallis H test, we compared individual feedback within the group that gives the same type of feedback. 
We found a significant difference ($H = 987.75, p < 0.0001$) within the group where participants were asked to give binary feedback, and a significant difference ($H =3461.77, p< 0.0001 $) within the group where participants were asked to give scalar feedback. 
To investigate the uniqueness of participants' feedback patterns, we performed a Wilcoxon Rank-Sum test for each participant and paired them with other participants within their groups (i.e. for each participant, we performed a Wilcoxon Rank-Sum test 44 times) and corrected the p-values with a step-down method using Holm-Sidak adjustments.
The results of the Wilcoxon Rank-Sum tests are shown in Figure \autoref{subfig:pttbp}. 
For  participants that gave binary feedback,  their feedback patterns on average had a significant difference with 25.58 participants (58.1\%).
For  participants that gave scalar feedback,  their feedback patterns on average had a significant difference with 37.13 participants (84.4\%).
That is, even correcting for the large number of tests, participants' feedback patterns were drawn from different distributions for most pairs of participants, supporting \textbf{H2}.



\begin{figure*}
    \centering
    \setlength{\abovecaptionskip}{0.2cm}
    \includegraphics[width = 0.95\linewidth]{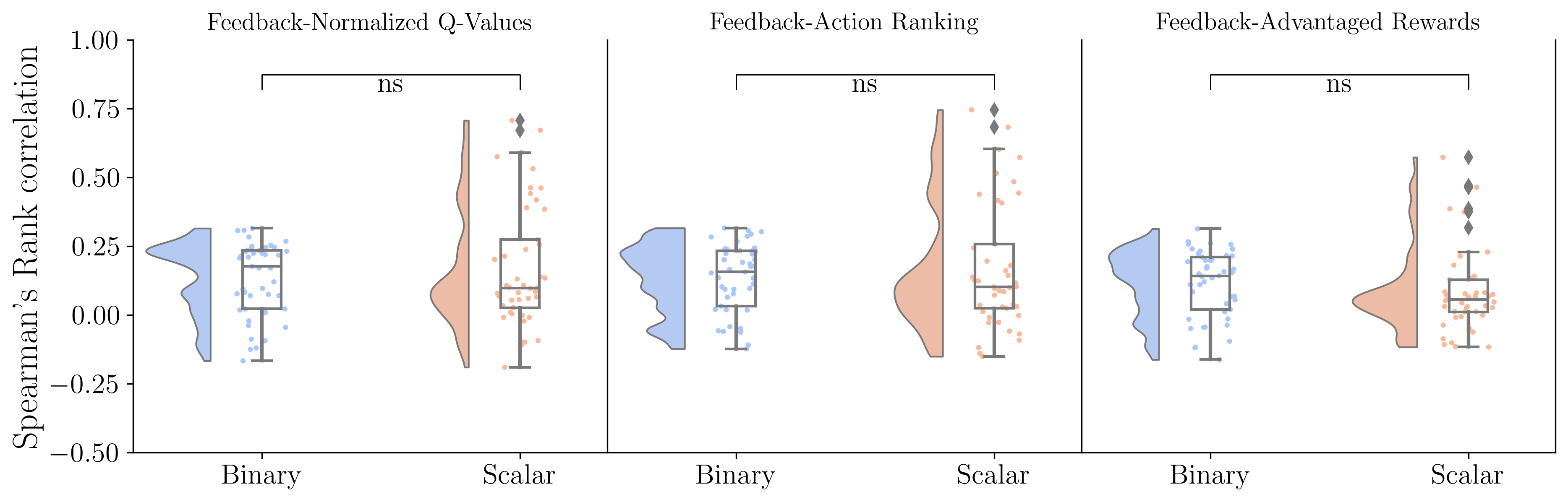}
    \caption{Correlation between feedback and rewards. From left to right, feedback-normalized Q-values (for TAMER), feedback-action ranking (for policy shaping), and feedback-advantaged rewards (for COACH).
    No significant differences had been found between binary feedback and scalar feedback in all three correlations. 
    }
    \label{fig:correlation}
     \vspace{-0.9em}
\end{figure*}
\subsection{Analysis of H3: How do scalar and binary feedback correlate with the learning algorithm's learning targets?}

To improve learning, Interactive RL uses human feedback to estimate internal elements of  RL algorithms \cite{arzate2020survey}; different Interactive RL algorithms correspond feedback with a different part of the learning problem.
We calculated the Spearman's Rank correlation between the feedback values and the training targets used in several key Interactive RL algorithms: advantaged rewards (COACH \cite{pmlr-v70-macglashan17a}), normalized Q-values (TAMER \cite{knox2008tamer}, and action rankings (Policy Shaping \cite{griffith2013policy}), using a standard MDP formulation with states $s \in S$, actions $a \in A$, and value functions $V(s)$ and $Q(s,a)$.
The advantaged rewards were computed by using the advantage function: $A(s,a) =Q(s,a) - V(s)$. The normalized Q-values were computed based on the equation: $r = \frac{Q(s,a)}{\sum_{a_i \in A} Q(s,a_i)}$. The action ranking assigned the rank $(0, \cdots, |A|-1)$ to each action $a$ based on the magnitude of $Q(s,a)$. 
The results are shown in \autoref{fig:correlation}.
Using Wilcoxon Rank-Sum test, no significance has been found in these results (feedback-normalized Q-Values $p =0.97$, feedback-action ranking $p = 0.75$, and feedback-advantage rewards $p = 0.81$),
so \textbf{H3} is not supported. 

\section{Stabilizing Teacher Assessment Dynamics}
\label{sec:steady}
The user study results suggest that scalar feedback is as good as binary feedback, but noise needs to be mitigated.
In previous work using scalar feedback,
scalar feedback is limited to a small range (e.g., COACH \cite{pmlr-v70-macglashan17a} is evaluated with scalar feedback with a range of $\{-1, 1, 4\}$), and must contain a negative range \cite{knox2008tamer, arakawa2018dqn, cederborg2015policy} to prevent algorithms from performing poorly.
To address these issues and to confirm that scalar feedback has advantages for learning, we present Stabilizing TEacher Assessment DYnamics (STEADY)
to reduce the noise in scalar feedback and extend the use of scalar feedback  by enabling binary-feedback-based methods to learn from scalar feedback.

\begin{figure}[t]
    \centering
    \includegraphics[width=0.9\linewidth]{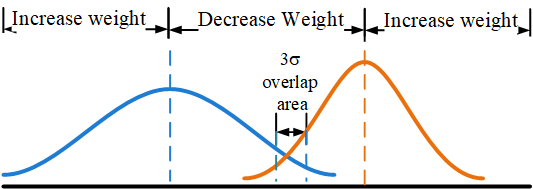}
    \caption{STEADY Visualization. STEADY redistributes scalar feedback into a preferable (orange) and a non-preferable (blue) distribution, and re-weights feedback. }
    \label{fig:cases}
        \vspace{-0.9em}
\end{figure}
\subsection{STEADY}
Our key intuitions are:
First, human evaluations often involve associating certain qualities with certain ranges. These are subject to variation,
and each range can be viewed as one distribution.
Second, since some robot behaviors are more preferred than others, feedback distributions can be separated into at least two classes, preferable (positive) and non-preferable (negative).
Based on the key intuitions,
STEADY re-constructs feedback distributions,
labels scalar feedback with binary labels,
and remaps the magnitude of the feedback in a more algorithmically useful way based on feedback statistics. 

\textbf{Initialization}
In an online learning setting,
the robot needs to learn from human feedback in real time and update its policy.
The feedback distributions thus also need to be updated after new feedback is added. 
Since there is no pre-labeled scalar feedback, we use a heuristic method to initialize the distributions.
For the first $k$ feedback, a midpoint method is used, which classifies the feedback above the average value as positive and otherwise negative.
In this work, we set $k=20$. 

\textbf{Distribution distance} The distance used is  Wasserstein distance \cite{kantorovich1960mathematical, ramdas2017wasserstein}.
The Wasserstein distance between the positive distributions $\phi^+$ and  the negative distribution $\phi^-$ is:
\begin{equation} 
\label{eq:wass dis}
\varpi (\phi^+, \phi^-)  = \inf_{\pi \in \Gamma (\phi^+, \phi^-)}\int_{\mathbb{R}\times\mathbb{R}}^{}\left | x-y \right | d\pi(x,y)
\end{equation}
where $\Gamma (\phi^+, \phi^-)$ is the set of (probability) distributions on $\mathbb{R}\times\mathbb{R}$ whose marginals are  $\phi^+$ and $\phi^-$.

\begin{algorithm}[b]
\caption{Stabilizing TEacher Assessment Dynamics (STEADY)} 
\label{alg:STEADY}
\begin{algorithmic}[1]

\State $\phi^+$, $\phi^-$ = initialize() 
\While {Human is still in loop}

\State Ask human feedback $\vec{f}$
\If {$\varpi(\phi^+ \cup \vec{f} , \phi^- ) >\varpi(\phi^+  , \phi^- \cup \vec{f} ) $ }
\State $\vec{f} \rightarrow \phi^+$, label $\vec{f}$ as \textit{positive}
\Else 
\State $\vec{f} \rightarrow \phi^-$, label $\vec{f}$ as \textit{negative}
\EndIf
\State \textit{$c_f$} = $conf(\vec{f},\phi^+,\phi^-)$
\If {$\vec{f}$ is in $3\sigma$ intervals of both $\phi^+$ and $\phi^-$ }
\State reduce\_overlap()
\EndIf
\EndWhile
\State return $c_f$ and binary label
\end{algorithmic}
\end{algorithm}

\begin{figure*}[htbp]
    \centering
    \includegraphics[width =  0.95\linewidth]{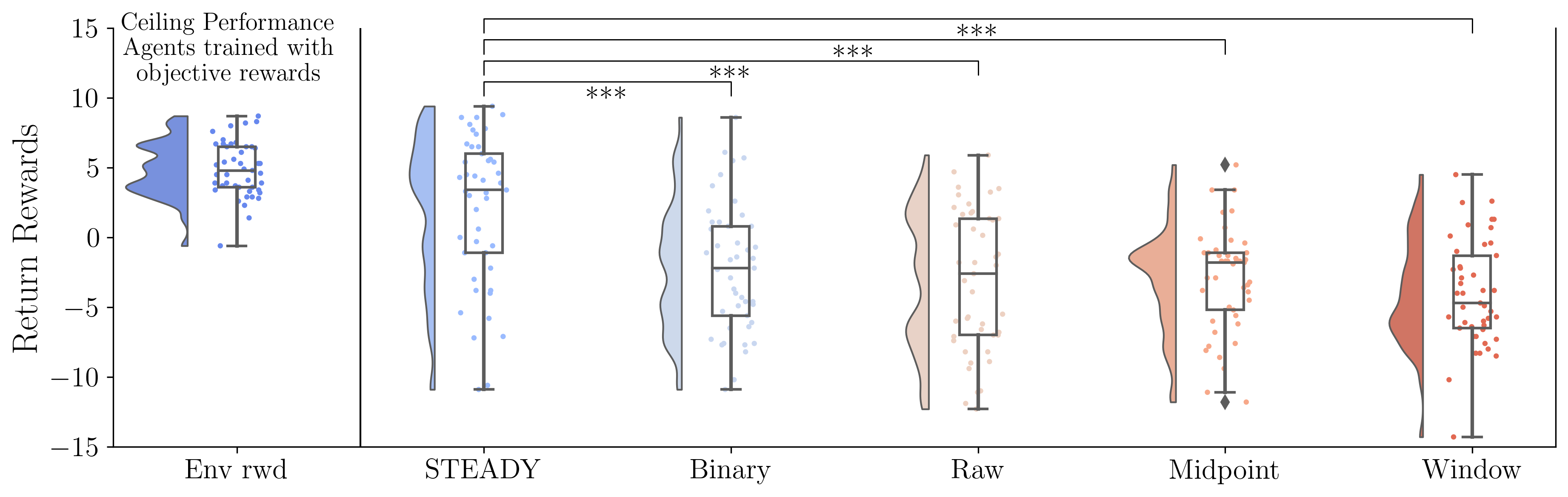}
    \caption{STEADY performance.  Each point represents a model's average performance over 10 runs, and each model is trained from one user's feedback. Models with \textit{Scalar feedback + STEADY} achieve significantly better performance than models with \textit{Binary feedback} and \textit{Raw Scalar feedback}.
    }
    \label{fig:steady performance}
        \vspace{-0.9em}
\end{figure*}

\textbf{Confidence Degree}
We use an inferred magnitude, which we refer to as the  \textit{confidence degree} to describe the intensity of feedback.
The \textit{confidence degree} can be used to improve learning by directly replacing raw scalar feedback with \textit{confidence degree} $\times$ \textit{binary label}. 
The \textit{confidence degree}
of feedback $\vec{f}$  is the degree to which $\vec{f}$  deviates from the center of its classified distribution, in the direction away from or towards the other distribution. 
Thus, there are two different cases for the \textit{confidence degree}, which are visualized in \autoref{fig:cases}.
For feedback $\vec{f}$ and its classified distribution $\phi$, the \textit{confidence degree} of \textit{increase weight} cases is given by:
\begin{equation}
\label{eq:conf1}
    \textit{conf}(\vec{f}) = 1 +\left | \int_{\mu (\phi^{'})}^{\mu(\phi)}\phi(x)dx \right |+  \left | \int_{\mu(\phi)}^{\vec{f}}\phi(x)dx \right |
\end{equation}
In \textit{decrease weight} cases, the \textit{confidence degree} is given by:
\begin{equation}
\label{eq:conf2}
    \textit{conf}(\vec{f}) =  1 - \left | \int_{\mu (\phi^{'})}^{\mu(\phi)}\phi(x)dx \right | + \left | \int_{\mu(\phi)}^{\vec{f}}\phi(x)dx \right |
\end{equation}
where $\mu$ is the mean of the distribution and $\phi^{'}$ is the non-classified distribution. 

\textbf{Overlap Reduction}
To differentiate the positive and negative distributions, we introduce a compensation mechanism.
If feedback $\vec{f}$ is in three-sigma areas of both distributions, STEADY pops the minimal feedback of the positive distribution and adds it into the negative distribution and vice versa unless $\vec{f}$ is minimal/maximal feedback.
This 
reduces the overlap area and differentiates the two distributions by moving outliers between the two distributions.

\textbf{STEADY} 
Algorithm \autoref{alg:STEADY} describes the overview of STEADY. 
After initialization, STEADY updates the distributions by adding new feedback to the distribution that maximizes the $\varpi (\phi^+, \phi^-)$.
We introduce an overlap reduction mechanism to differentiate the distributions, and provide improved magnitudes (effectively a measure of confidence in the label) from the constructed distributions.

\textbf{Multiple Distribution Extension}
We use STEADY in a  two-distributional case, but STEADY can be easily extended to $m$-distributional scenarios by selecting $m$-th percentiles instead of maximum and minimum and maximizing the distance described in \autoref{eq:2} instead of \autoref{eq:wass dis}.
The distance among $m$ distributions is:
\begin{equation}
    \label{eq:2}
    \varpi_\phi = \sum_{\phi_u,\phi_v,\phi_q \in \phi}^{} \varpi (\phi_u,\phi_v) + \varpi(\phi_v,\phi_q)
\end{equation}
where $\phi_u$ is adjacent to $\phi_v$, and $\phi_v$ is adjacent to $\phi_q$.

\subsection{Scalar Feedback with STEADY} 

We validated that STEADY enables learning from scalar feedback by training the robot to perform the button-pressing task using collected human feedback.
We use TAMER\cite{knox2008tamer} as our learning algorithm.
Human feedback is used as reward signal. 
Since the robot did not take any environmental rewards as input and had been initialized to be the same,
the only factor that impacts learning is human feedback.

\subsubsection{Baselines}
We have four baselines to compare the performance with \textit{scalar feedback + STEADY}:
Binary feedback; 
 Raw scalar feedback;
 Scalar feedback with a midpoint classifier; 
 Scalar feedback with a sliding window classifier. 
Additionally, we included the environmental reward (the same used to train the oracle) to show the ceiling performance. 
Raw scalar feedback was pre-processed by subtracting the midpoint value of the range (i.e. five) since TAMER performs poorly with scalar feedback without a negative range.
The Midpoint Classifier converts scalar feedback to binary by converting a scalar feedback value greater than five to positive and less or equal to five to negative.
Using the sliding window method, feedback higher than the mean of received feedback within a window is treated as positive feedback, while feedback less or equal to the mean is treated as negative feedback.  We used a window size of 20, which was empirically derived from a series of experiments with window sizes from 1 to 200.


\subsubsection{Trained Models}
For each baseline, we trained 45 models.
Each model is a TAMER agent trained by one participant's feedback. 
Thus, there are $5 \times 45 = 225$ models in total.
We initialized the models to the same state (completely untrained).
Each transition (state-action-feedback tuple)  is only learned once for each model and fed to the models in the same order. 
After finishing training, we applied the trained models to the robot and performed the button-pressing task.
For each model, we ran it 10 times on the robot to evaluate real-world performance while reducing the  random noise in the results.

\subsubsection{Model Performance}
We use returned environmental rewards to measure the performance of the models.
The maximum return reward can be 14, and the minimum can be -50; because the models were tested on the real robot, randomness in the results comes from noise in robot movement.
The performance is shown in  \autoref{fig:steady performance}.
The Kruskal-Wallis H test shows that the choice of the baselines has a significant impact on the trained agents' performance ($H = 105.04, p < 0.0001 $). 
Post-hoc analysis with the Wilcoxon Rank-Sum test shows that the models trained with STEADY have
a $4.22$ higher average return rewards than the models trained with the binary feedback ($statistic = 178.5, p < 0.0003 $),
a $5.12$ higher average return rewards than the models trained with the raw scalar feedback ($statistic = 63.0, p < 0.0001 $),
$4.99$ higher than the models trained with \textit{midpoint} classifier ($statistic = 82.0, p < 0.0001 $), 
and $6.07$ higher than the models trained with \textit{sliding window} ($statistic = 37.0, p < 0.0001 $).
STEADY enables learning from scalar feedback and 
models with STEADY have a  higher average return reward than all baselines using human feedback including binary feedback.

\section{Discussion \& Conclusion}
The results above demonstrate that scalar feedback is not inherently worse than binary feedback as a teaching signal even without stabilization. Scalar feedback does change more than binary feedback, but tolerating small variations removes the effect.
Furthermore, the average time per assignment provided by AMT
suggests that for crowdworkers, giving scalar feedback does not put an extra time burden  on participants.
Our results show that human feedback patterns in both scalar and binary feedback, but especially scalar feedback patterns, tend to be different in statistically distinguishable ways. 
This explains why learning from multiple participants is likely to be difficult and suggests that
learning agents may be able to
differentiate between teachers based on their individual feedback patterns. 
In addition, our results show that scalar feedback with stabilization can be leveraged to improve learning.  That is,
scalar feedback with STEADY achieves significantly better performance than binary feedback and other baseline methods.
This conclusion should be able to extend to more complex tasks and other algorithms that have been shown to work with binary feedback since STEADY does not require changing any part of the learning algorithms and only filters human feedback.

One limitation of our work is that the experiments were relatively short-term and only repeated each clip once.
We are aware that human teachers were learning and adapting during teaching in the short term, but we did not address long-term feedback dynamics.
Repeating the robot behaviors only once was necessary to prevent participants from becoming aware that clips were repeating, but this limits our ability to understand the full range of possible changes in feedback dynamics.
Future work could investigate longer-term changes in feedback and new methods for stabilizing dynamics over long periods of time.

Overall, our work suggests that HRI and interactive RL practitioners should give scalar feedback more attention, including developing new learning algorithms to utilize the additional information, finding new methods to collect more accurate scalar feedback, and considering using scalar feedback and binary feedback together. STEADY provides a bridge between algorithms that are designed and validated with binary feedback and the next generation of algorithms that can appropriately leverage scalar feedback.  Future studies can build on these results and more fully leverage the scalar information that is available from human teachers.

\balance
\bibliographystyle{IEEEtran}
\bibliography{IEEEexample}

\end{document}